\def\year{2022}\relax
\documentclass[letterpaper]{article} 
\usepackage{aaai22}  
\usepackage{times}  
\usepackage{helvet}  
\usepackage{courier}  
\usepackage[hyphens]{url}  
\usepackage{graphicx} 
\usepackage{natbib}  
\usepackage{caption} 

\usepackage{subcaption}
\usepackage{amsfonts, amssymb}
\usepackage{multirow}
\usepackage{booktabs}
\usepackage{pifont}
\usepackage{pdfpages}

\DeclareCaptionStyle{ruled}{labelfont=normalfont,labelsep=colon,strut=off} 
\usepackage{algorithm}
\usepackage{algorithmic}

\usepackage{newfloat}
\usepackage{listings}
\floatstyle{ruled}
\newfloat{listing}{tb}{lst}{}
\floatname{listing}{Listing}
\title{Stage Conscious Attention Network (SCAN) :\\ A Demonstration-Conditioned Policy for Few-Shot Imitation}
\author{
    Jia-Fong Yeh\textsuperscript{\rm 1}\equalcontrib, 
    Chi-Ming Chung\textsuperscript{\rm 1}\equalcontrib, 
    Hung-Ting Su\textsuperscript{\rm 1},
    Yi-Ting Chen\textsuperscript{\rm 2},
    Winston H. Hsu\textsuperscript{\rm 1, 3}
}
\affiliations{
   \textsuperscript{\rm 1}National Taiwan University\\
   \textsuperscript{\rm 2}National Yang Ming Chiao Tung University\\
   \textsuperscript{\rm 3}Mobile Drive Technology
}

\usepackage{bibentry}

\begin{document}

\maketitle

\begin{abstract}
In few-shot imitation learning (FSIL), using behavioral cloning (BC) to solve \textbf{unseen tasks} with few expert demonstrations becomes a popular research direction. The following capabilities are essential in robotics applications: (1) Behaving in compound tasks that contain multiple stages. (2) Retrieving knowledge from few length-variant and misalignment demonstrations. (3) Learning from a different expert. No previous work can achieve these abilities at the same time. In this work, we conduct FSIL problem under the union of above settings and introduce a novel stage conscious attention network (SCAN) to retrieve knowledge from few demonstrations simultaneously. SCAN uses an attention module to identify each stage in length-variant demonstrations.  Moreover, it is designed under demonstration-conditioned policy that learns the relationship between experts and agents. Experiment results show that SCAN can learn from different experts \textbf{without fine-tuning} and outperform baselines in complicated compound tasks with explainable visualization.


\end{abstract}

\section{Introduction}
\begin{figure}[t]
    \centering
    \includegraphics[width=0.9\columnwidth]{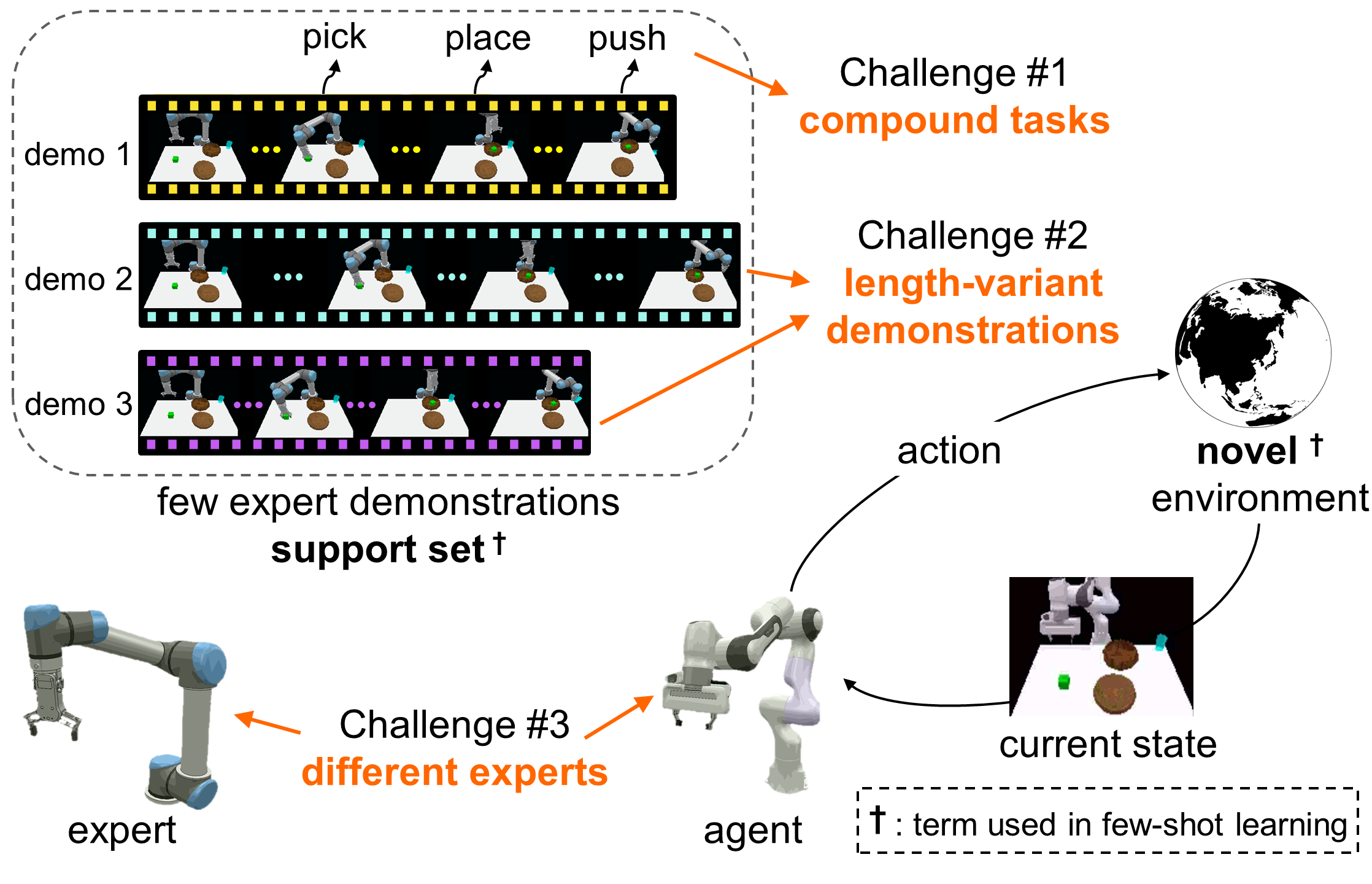}
    \caption{\textbf{Schema of few-shot imitation learning (FSIL).} In our FSIL problem, models need to solve the task in \textbf{novel} environment that is unseen during training. Few demonstrations are given to let models imitate. There are three challenges in our FSIL setting, (1) We conduct FSIL on compound tasks which contain multiple stages. (2) Demonstrations are length-variant. Each stage may locate at different timestamps. (3) Models need to learn the behavior from a different type of expert.  None of previous works can solve these challenges concurrently. Moreover, learning from length-variant sequences is non-trivial, making our FSIL a challenging yet practical problem.}
    \label{fig:fsil}
\end{figure}

\noindent Humans can learn to perform \textbf{unseen compound tasks} from different experts with few nonidentical demonstrations. The number of researches on few-shot imitation learning (FSIL) increases rapidly to verify whether the machine also has this ability. The above challenging settings of FSIL problem are illustrated in Figure \ref{fig:fsil}, and most of existing works only solve them partially. To overcome the complexity of environment and concomitant complicated training process,  behavioral cloning (BC) is leveraged to mimics the experts. Previous works \cite{finn17a, Duan2017, Yu2018, Yu2019, dasari2020transformers, Bonardi2020} only support one demonstration at once, which limits the capability of models. We argue that retrieving knowledge from few demonstrations simultaneously can break through the limitations and lead to better performance when conducting FSIL under these challenging settings.

Meta-learning based methods \cite{finn17a, Yu2018, Yu2019} learn a meta-policy $\pi(a \mid s)$ that takes state $s$ from current playout $p$ and outputs an action $a$ via BC. Before testing, the meta-trained policy uses expert demonstrations to adapt meta-parameters. Then, they update the policy several times based on the number of demonstrations. The main drawbacks of these methods are twofold: (1) fine-tuning is required, and (2) they need a learnable loss \cite{Yu2019} to map expert to agent explicitly. 

To tackle these drawbacks, demonstration-conditioned (DC) based methods
\cite{Duan2017, James2018,  Bonardi2020, Shao2020ObjectDO, dasari2020transformers, dance21} apply a policy $\pi(a\mid s, \mathcal{D})$ which predict actions conditioned on both states and demonstrations $\mathcal{D}$.  Fine-tuning is optional for DC methods, because they are expected to behave by observing demonstrations. They implicitly learn the mapping since they have the information in $\mathcal{D}$ concurrently when generating $a$, even if experts are different from the agent. The primary objective of DC policies is to encode few demonstrations into a representative embedding.  However, it is difficult to encode length-variant demonstrations. Hence, most DC works \cite{Duan2017, Shao2020ObjectDO,dasari2020transformers} only contains one demonstration in $\mathcal{D}$ and claim themselves one-shot imitation methods.

To handle demonstrations with variant lengths, some DC works apply task-embed techniques. The authors \cite{James2018, Bonardi2020} concatenate the first and last frames of each demonstration and average their features to generate task-embedding (alias sentence in their paper).  Nevertheless, this method does not consider the importance of temporal transitions that are essential for policy learning and cannot provide efficient information in the case of long demonstrations given. Therefore, \citeauthor{dance21}  \shortcite{dance21} introduces a transformer-based network that considers both temporal and cross-demonstration information using attention mechanisms. They generate a task embedding by averaging the attention output of few demonstrations at each timestamp. The premise of this method is that frames at the same timestamp of each demonstration have similar knowledge. However, due to different initial states, the operating time of each stage could easily vary and result in temporal misalignment. Therefore, the model might be confused by the frame mixed with two 
distinct stages and degrade the performance.

We aim to design an attention mechanism that can identify important frames at different timestamp. Meanwhile, the attention mechanism should detect stages in compound tasks. A compound task containing multiple stages often appears in robotics problems.  When solving compound tasks in FSIL problem, the policy needs to learn both perception and path planning. This makes solving compound tasks challenging. \citeauthor{Yu2019} \shortcite{Yu2019} leverages an additional phase predictor to split the compound tasks, and then the policy only needs to adapt to each stage. The disadvantage is that the number of stages needs to be known in advance.

In this work, we propose a novel stage conscious attention network (SCAN). SCAN takes both demonstrations and playouts as input to learn the mapping due to the characteristic of DC method. Furthermore, SCAN applies novel \textbf{stage conscious attention} to let each playout frame has its attention score to each frame in the demonstrations. The frame features of demonstrations are weighted by attention scores to produce the same shape contexts.  We then average them to generate the informative task-embedding. With stage conscious attention, SCAN can retrieves knowledge from few demonstrations simultaneously. Experimental results show that SCAN has a significant performance improvement compared to baselines with explainable visualizations. The overall contributions are summarized as follows:
\begin{itemize}
    \item Our work is the first DC method that solves FSIL problem under the settings of compound task, length-variant demonstrations, and learning from a different expert.  
    \item The novel stage conscious attention detects important frames of misalignment stages and is robust to length-variant demonstrations.
    \item Extensive experiment results express proposed SCAN is powerful, and explainable visualization also proves the effectiveness of novel stage conscious attention. 
\end{itemize}

\section{Related Work}
\paragraph{Few-shot Learning.} Few-shot learning (FSL) has become popular since collecting a huge amount of labeled data is difficult in most research problems. The objective of FSL is to infer the unlabeled data (query set) by leveraging few labeled data (support set). FSL is first studied on image classification.  There are many influential and well-known metric-learning-based FSL methods, such as Matching network \cite{Vinyals16MN}, Prototypical network \cite{Snell17PN}, and Relation Network \cite{Sung18RN}. They try to learn the relationship between support and query sets rather than inferring the unlabeled data directly. Moreover, optimization-based method \cite{Finn17MAML} seeks a meta-parameter set that can quickly adapt to unseen tasks. Nowadays, FSL has been extended to many research fields. For instance, image segmentation \cite{SSegment20}, object detection \cite{FSOD21}, and imitation learning \cite{FSIL2020}. Our work aims to develop a method that can learn unseen compound tasks with few-shot length-variant demonstrations from a different expert.


\paragraph{Few-shot Imitation Using RL/IRL.} Reinforcement learning (RL) methods assume the reward function of environments is known.  But, it is difficult to design reward functions that give precise feedback to policies in real-world problems. Therefore, inverse RL (IRL) \cite{Ng99policyinvariance} infers a reward function from few expert demonstrations. Then, IRL policy can be trained by interacting with the environment under the inferred reward function. In addition, modern IRL methods \cite{Ho16, reddy2020sqil} are usually GAN-like \cite{goodfellow2014generative}. They assign a high reward to the states from demonstrations but a low reward to the states from collected agent samples. Since these methods only use rewards of states to train their model, they can handle the demonstrations without actions, which differs from BC. However, both RL and IRL methods need to interact with the target environment to train the policy. In other words, the learned reward function (IRL) or the well-trained policy (RL) do not be applicable to novel environments. In our FSIL problem, policies can only use demonstrations to fine-tune but cannot interact with the novel environment before performance evaluation. Thus, most RL and IRL methods are not available in our work.

A recent work \cite{dance21}, named demonstration-conditioned RL (DCRL), overcomes the limitation. DCRL requires interactions with environments in training. But, it solves FSIL tasks without fine-tuning in testing. Because DCRL is a DC policy method, it needs expert demonstrations from training environments to achieve fine-tuning-free. They store the tuples of (playout history, rewards, demonstrations) into the replay buffer to train the policy. The policy is a transformer-based architecture that its encoder generates task embeddings using cross-demonstration attention and the decoder predicts the actions. Moreover, the concept of "demonstration conditioned" echoes our motivations. But, designing all reward functions in training environments is quite time-consuming. Thus, we only compare SCAN with BC-based methods.
 
\paragraph{Compound Task.} A compound task consists of multi-stage subtasks. The main challenge of compound tasks is that there is no signal (label) in demonstrations to identify each subtask when testing. Nevertheless, the policies for distinct subtasks are pretty different. Hence, it becomes impractical to use a single policy to solve compound tasks. 

An intuitive solution is to link the relationship between subtask and \textit{primitive}. The term \textit{primitive}, which comes from the robotics field \cite{FLASH2005660, MANSCHITZ201597}, represents single elementary movements and is widely used in compound task problems. To be specific, each subtask may corresponds to a single robot motion (primitive), such as pushing and grasping. Therefore, previous RL \cite{zeng2018learning, marzari2021hierarchical} and IL works \cite{Yu2019, lee2019composing, pmlr-v119-lee20f} assume these primitives are known before testing since the end effector (gripper) are only capable for these primitives. Then, they develop a hierarchical structure that trains policies for each primitive separately and use a high-level control network to decide which policy should be executed with the current observation. Phase predictors \cite{Yu2019} or task transition models \cite{lee2019composing,pmlr-v119-lee20f} are proposed to identify whether a primitive is done. Unfortunately, most of these components cannot be trained from scratch with actor networks due to the hierarchical architecture. Instead of recognizing primitives with an additional model, SCAN detects which primitive the current observation locates by an attention module. Unlike above methods, SCAN does not need to know the primitives beforehand, and all its components can be trained end-to-end.

\section{Few-Shot Imitation Learning (FSIL)}
To emphasize, we treat FSIL problem as imitation learning under FSL setting. In previous works, a policy is (trained or) tested over several tasks that are essentially the same but with different objects in the environments. Thus, we use the base environments $E^{b}$ and novel environments $E^{n}$ in problem statement. Notations are listed in Table \ref{table:notation}. 

\subsection{Problem Statement} 
A \textit{few-shot imitation learning (FSIL)} problem is given by a base environment set $E^{b}$ and a novel environment set $E^{n}$, where $E^{b} \cap E^{n} = \emptyset$. Each environment $e^{*}$ in $E^{b}$ or $E^{n}$ contains a set of expert demonstrations (support set) $\mathcal{D}^{*}$ and a set of agent playouts (query set) $\mathcal{P}^{*}$.
In addition, samples in $\mathcal{D}^{*}$ or $\mathcal{P}^{*}$ are in the same environment $e^{*}$ with distinct initial/end states. Moreover, the expert could be humans or other robot arms compared to the agent in playouts. 

A policy $\pi_{\theta}$ with parameter $\theta$ is meta-trained in $e^{b}$ from base environments $E^{b}$ and meta-tested in $e^{n}$ from novel environments $E^{n}$. The policy $\pi_{\theta}$ needs to generate an action $a$ when receiving a current state $s$ from the playout $p^{n} \in \mathcal{P}^{n}$. Then, the success rates of playout executions in all $e^{n}$ are used as performance evaluation. Thus, the playout samples in $e^{n}$ only contain the initial state $s_{0}$, and the following states are provided according to the action taken by the policy. At last, the objective of FSIL problem is to maximize the performance expectation of the policy where the expectation is taken over $e^{n} \in E^{n}$. Furthermore, the policy can only fine-tune its parameters using the demonstrations $\mathcal{D}^{n}$ in $e^{n}$(if needed), which means no interaction with novel environments are allowed before performance evaluation. 

\begin{figure*}[ht]
    \centering
    \includegraphics[width=.75\linewidth]{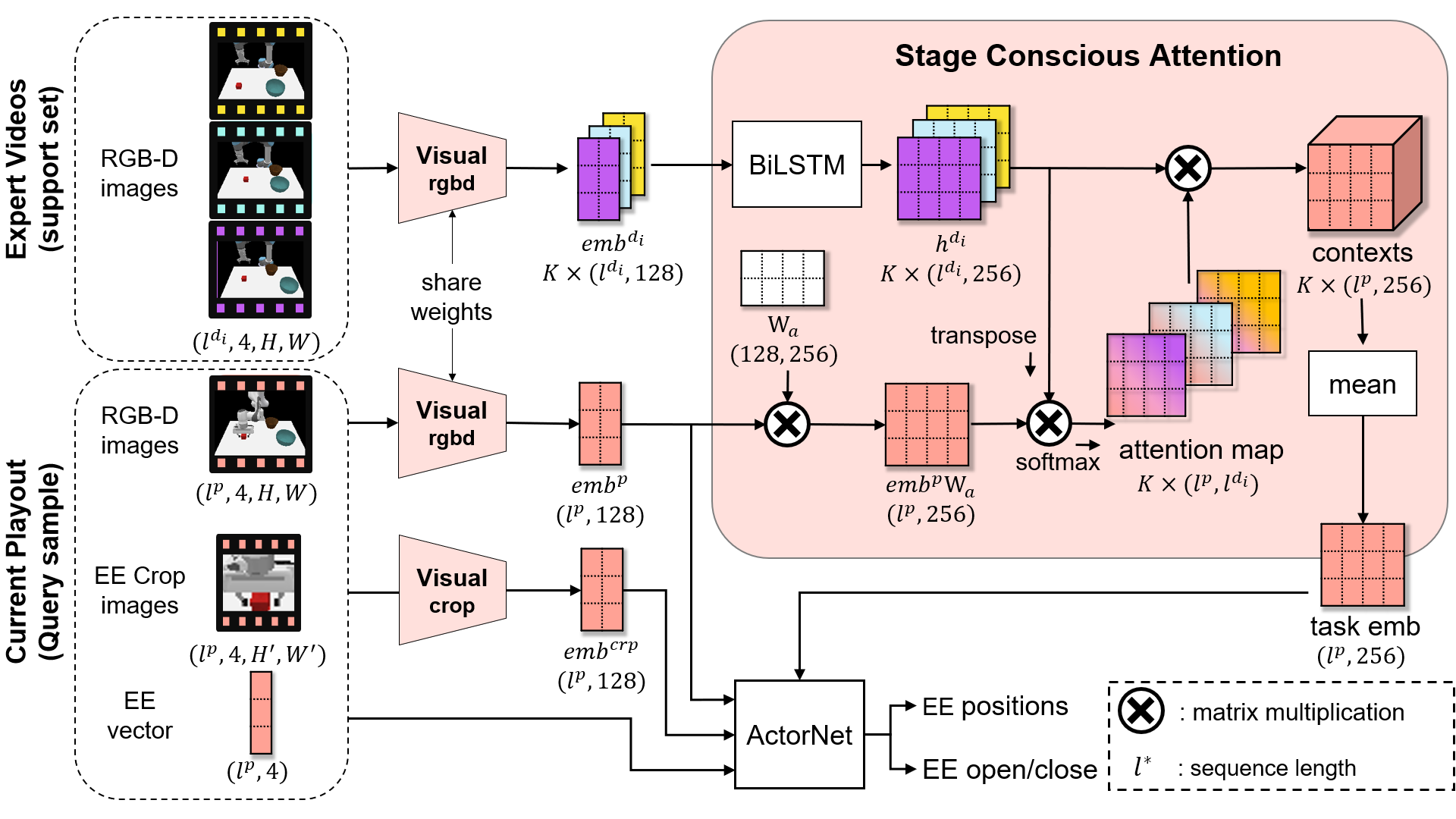}
    \caption{\textbf{Architecture of stage conscious attention network (SCAN).} Given a set of length-variant demonstrations (images only) and current playout data (images, end-effector (EE) cropped images, EE vector), SCAN uses a shared-weight $Visual_{rgbd}$ to extract feature embeddings (FEs) $emb^{d_{i}}$ from demonstrations and $emb^{p}$ from current playout. Meanwhile, another $Visual_{crop}$ generates FEs of EE cropped images  $emb^{crp}$. Then, $emb^{d_{i}}$ are passed into a bidirectional LSTM (BiLSTM, encoder) to generate $h^{d_{i}}$ that contains time information. In addition, the attention maps are the results of matrix multiplications of $emb^{p}$ and a learnable matrix $W_{a}$ and each $h^{d_{i}}$. 
    Afterwards, attention maps matrix multiplies by each $h^{d_{i}}$ to get contexts. Until now, length-various demonstrations are projected to same shape contexts. The task embedding (mean of contexts) contains the knowledge that each playout state focuses on. We named this process as \textbf{stage conscious attention (SCA)}. At last, $emb^{p}$, $emb^{crp}$, ee vector and task-embeddings are fed into ActorNet (decoder) for predicting actions.}
    \label{fig:model}
\end{figure*}

\begin{figure}[t]
    \centering
    \includegraphics[width=.80\columnwidth]{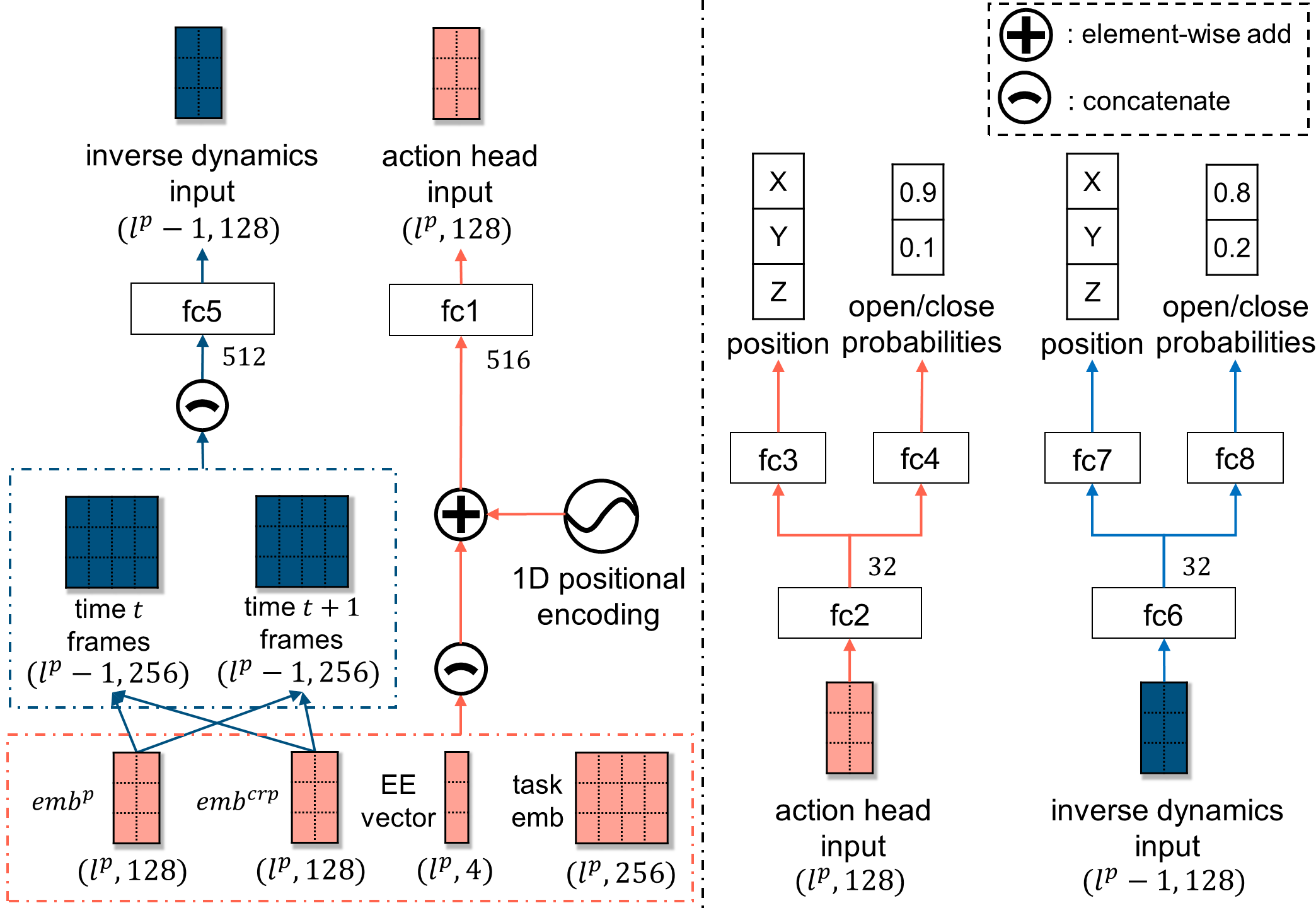}
    \caption{\textbf{Architecture of ActorNet.} In ActorNet, an action head concatenates four input embeddings and adds a 1D positional encoding. Following by the several dense layers,the action head computes positions and probabilities of open/close control. In the meantime, an inverse dynamics model concatenates time $t$ and $t+1$ $emb^{p}$ and $emb^{crp}$ as inputs and predict the actions using a similar architecture of action head. The inverse dynamics model helps the action head know the precise actions that cause changes between frames.}
    \label{fig:actor}
\end{figure}

\begin{figure}[!t]
    \centering
    \begin{subfigure}[b]{.49\columnwidth}
    \centering
        \includegraphics[width=0.5\textwidth]{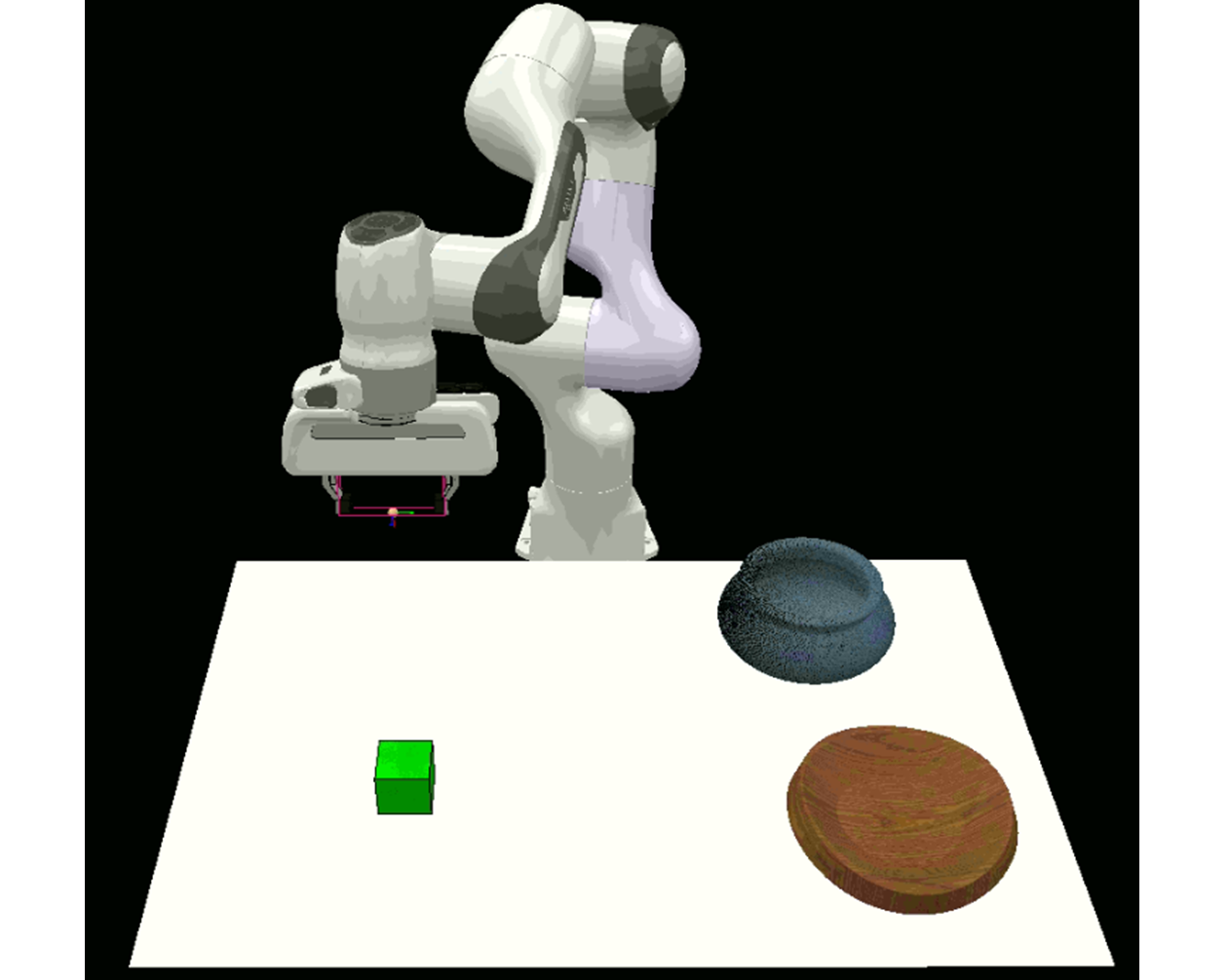}
    \caption{\textit{pick \& place (PP)}}
    \end{subfigure}
    \begin{subfigure}[b]{.49\columnwidth}
    \centering
        \includegraphics[width=0.5\textwidth]{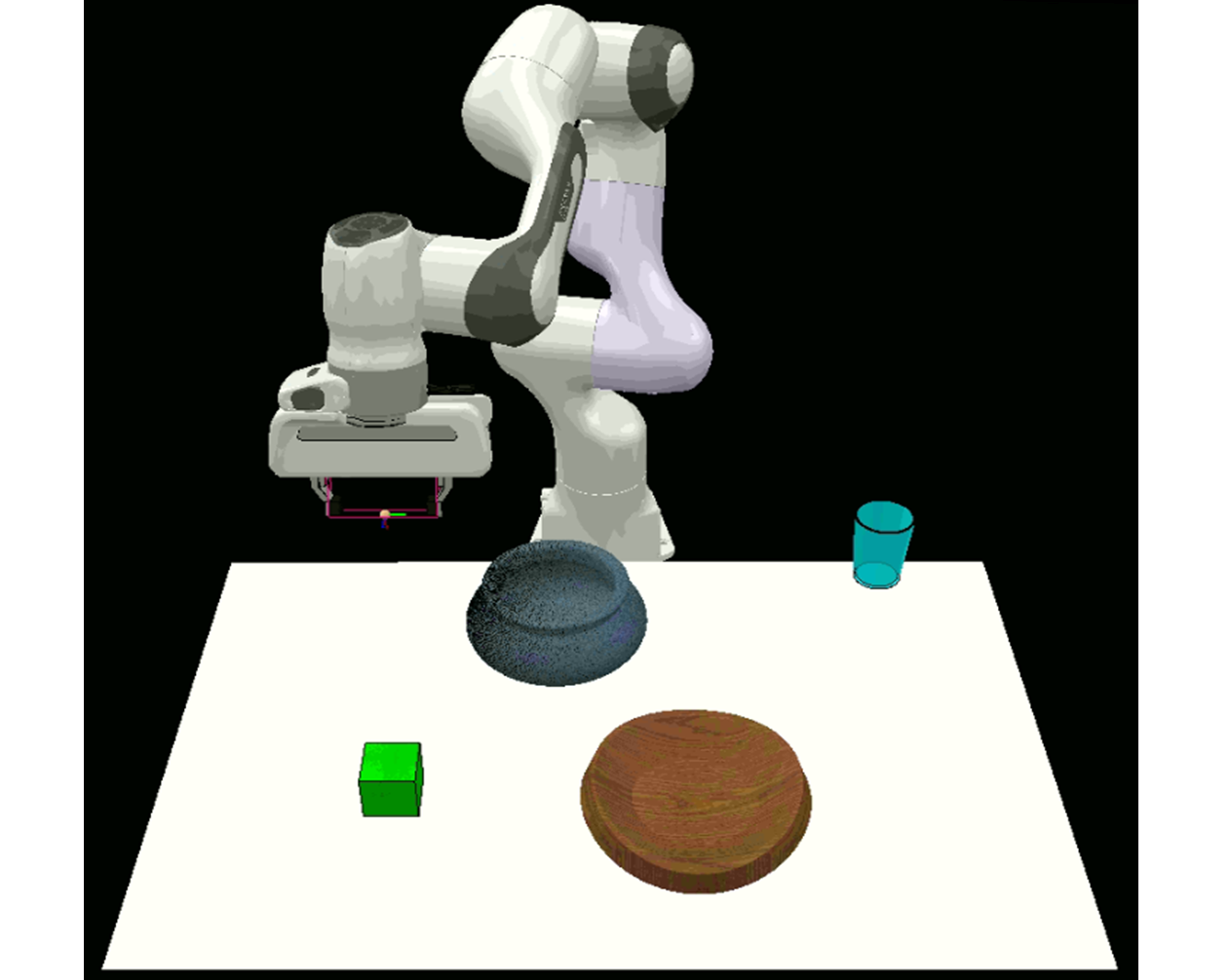}
    \caption{\textit{pick \& place \& push (PPP)}}
    \end{subfigure}
    \caption{Environments of \textit{PP} and \textit{PPP task} . \textit{PP} and \textit{PPP task} are compound tasks that contains two and three stages, respectively. Objects are unseen during training.}
    \label{fig:env}
\end{figure}


\begin{table}[!t]

\centering
\begin{tabular}{|ll|}
\hline
notation                                                     & meaning \\
\hline
$f, f^{crp}$                                                 & frame, crop frame \\
$\vec v$                                                     & vector \\
$s := (f, f^{crp}, \vec v)$                                     & state \\
$a$                                                          & action \\
$\mathbb{F} := [f_{0},f_{1}, ..., f_{t}]$                   & sequence of frames \\
$\mathbb{F}^{crp} := [f_{0}^{crp},f_{1}^{crp}, ..., f_{t}^{crp}]$ & sequence of crop frames \\
$\vec {\mathbb{V}} := [{\vec v}_{0},{\vec v}_{1}, ..., {\vec v}_{t}]$  & sequence of vector \\
$\mathbb{S} := [s_{0},s_{1}, ..., s_{t}]$                   & sequence of state \\
$\mathbb{A} := [a_{0},a_{1}, ..., a_{t}]$                   & sequence of action \\
$p^{*} := (\mathbb{S}, \mathbb{A})^{*}$                          & playout \\
$d^{*} := (\mathbb{F}, \mathbb{A})^{*}$                          & demonstration \\
$\mathcal{P}^{*}, \mathcal{D}^{*}$ & set of $p^{*}, d^{*}$ \\
$e^{*}$                                                      & environment\\
$E^{*}$                                                      & set of $e^{*}$\\
$^{*}\in\{b,n\}$  & $b$: base, $n$: novel \\ 
\hline
\end{tabular}
\caption{Defined notations}
\label{table:notation}
\end{table}

\section{Methodology}
We introduce the stage conscious attention networks (SCAN) in details. SCAN needs following inputs: $K$ expert demonstrations (without actions) $\mathcal{F}^{D} = \{\mathbb{F}^{d_{i}} \mid i\in [0, K] \}$ and playout states that contains frames $\mathbb{F}^{p}$, end-effector (EE) cropped frames  $\mathbb{F}^{crp}$, and EE vector $\vec {\mathbb{V}}^{p}$ (position and open amount). These inputs are widely used in FSIL works. Moreover, we apply inverse kinematics to compute joint parameters of the agent. Therefore, SCAN only needs to compute two outputs for each playout state: target positions (x, y, z in continuous space) and the probabilities for EE open/closing. SCAN is composed of three main components, including visual heads, stage conscious attention, and an ActorNet. We describe these components in following paragraphs, and the overall architecture of SCAN is shown in Figure \ref{fig:model}. 

\paragraph{Visual Head.} The objective of visual head is to retrieve meaningful embeddings of RGB-D frames either from $\mathcal{F}^{D}$ or $\mathbb{F}^{p}$. We leverage two resnet18 to extract RGB images and depth images separately, inspired by \cite{Shao2020ObjectDO}. Moreover, since SCAN does not have object detection components, we insert a modified self-attention module at the head and tail of resnet18. The self-attention module is originally proposed in \cite{zhang2019selfattention}, we make the output dimension of the query-layer and key-layer the same as input dimensions and add the channel-wise softmax layer behind them. Detailed architecture is shown in supplementary.

Then, given a 4D frame inputs (sequence length $l^{p}$ for  $\mathbb{F}^{p}$ or $l^{d_{i}}$ for $\mathbb{F}^{d_{i}}$, channel $C$, height $H$, width $W$), the visual head splits it into RGB and depth images and use the corresponding resnet18 to generate feature embeddings (FEs). These two FEs are concatenated and passed over a dense layer to obtain the output with shape ($l^{p}$ or $l^{d_{i}}$, 128). As shown in Figure \ref{fig:model}, a shared-weights $Visual_{rgbd}$ extracts $emb^{p}$ from $\mathbb{F}^{p}$ and each $emb^{d_{i}}$ from $\mathbb{F}^{d_{i}}$. Another $Visual_{crop}$ extracts $emb^{crp}$ from $\mathbb{F}^{crp}$.

\paragraph{Stage Conscious Attention (SCA).} We assume that each frames in each demonstration $\mathbb{F}^{d_{i}}$ (support sample) has a different importance to each frame in playout $\mathbb{F}^{p}$ (query sample). Unlike the cross-demonstration attention (temporal-awareness attention) in \cite{dance21},  
SCAN leverages a stage conscious attention that lets each frame of current playout has its own interpretation of each demonstration. The overall structure is shown in Fig. \ref{fig:model}.

We feed each $emb^{d_{i}}$ into a bidirectional LSTM (BiLSTM) to obtain $ h^{d_{i}}$ that contains \textbf{temporal information}. Then, the general form of global attention motivated by \cite{luong2015effective} is applied to generate an attention map $attn^{p}_{d_{i}}$ between $h^{d_{i}}$ and FEs of playout $emb^{p}$. To be specific,
$attn^{p}_{d_{i}}$ is the result of matrix multiplied of $emb^{p}$ and a learnable matrix $W_a$ and $ h^{d_{i}}$. 
The $W_a$ makes sure the $emb^{p}$ has a same latent size of $ h^{d_{i}}$, which is useful when dealing with two different shape matrix. Then, a $softmax$ layer makes each row (attention scores at each frame in the demonstration $d_{i}$ given by each frame in current playout) of attention map are summed to 1. Next, the context $c^{d_{i}}$ is the result of matrix multiplication of $attn^{p}_{d_{i}}$ and each $h^{d_{i}}$. At last, the task-embedding $emb^{task}$ is the mean of $K$ contexts $c^{d_{i}}$. 

\[ attn_{d_{i}}^{p} = softmax(emb^{p}W_{a}(h^{d_{i}})^{T}) \]
\[ c^{d_{i}} = attn_{d_{i}}^{p} \cdot h^{d_{i}}\]


\paragraph{ActorNet.} ActorNet predicts actions in continuous space rather than discrete spaces. It allows the agent to perform more precise actions. The overall picture of ActorNet is shown in Figure \ref{fig:actor}.  In ActorNet, there are two components, action head and inverse dynamics model. The action head concatenates four inputs ($emb^{p}$, $emb^{crp}$, EE vector $\vec {\mathbb{V}}^{p}$, $emb^{task}$) and add 1D positional encoding $PE_{1D}$ to provide auxiliary time information. Then the action head predicts target positions and probability of open/close for each state in current playout (history), we denote the output as $out_{act}$. 
\[ in_{act} = PE_{1D}([emb^{p}; emb^{crp}; \vec {\mathbb{V}}^{p}; emb^{task}]) \]

The $out_{act}$ is a vector of pairs $[\vec{\mu}_{t}, g_{t} ]$ for the time $t$ state in the playout history. The $\vec{\mu}_{t} = [\mu_{x}, \mu_{y}, \mu_{z}]$ is a vector which contains means of three uni-variate Gaussian distributions that generate positions at time $t$. We use an additional learnable vector $\vec{\sigma}$ for the standard deviation of the distributions. Besides, the $g_{t}$ is the probability of open/close control.

Afterwards, the inverse dynamic model concatenates time $t$ $emb^{p}$, $emb^{crp}$ and time $t+1$ $emb^{p}$, $emb^{crp}$ as input $in_{inv}$, and use the similar architecture of action head to predict actions $out_{inv}$ for each state in playout history (except for the latest frame). The inverse dynamic model aims to help the action head know how actions cause frame changes.

\[ in_{inv} = [[emb^{p}_{t}, emb^{crp}_{t}],[emb^{p}_{t+1}, emb^{crp}_{t+1}]]\]
To train SCAN, we use negative-log-likelihood (NLL) as loss functions. $L_{*}^{pos}$ calculates the NLL loss for the output $out_{*}$ of the predicted positions. Besides, $*$ could be $act$ or $inv$. And the $\vec{a} = [a_{x}, a_{y}, a_{z}]$ is the vector of labeled actions.
\[
L_{*}^{pos} = -\frac{1}{K}\sum_{i=1}^{K}\sum_{j \in [x, y, z]}\frac{1}{2}\ln(2\pi)+\ln(\vec{\sigma}_{j})+\frac{( a_{j}-\mu_{j})^{2}}{2\sigma_{j}^{2}}
\]

We use the following loss for open/close control. The $g^{label}\in\{0,1\}$ is true probability of open the gripper. And, $*$ indicates action head $act$ or inverse dynamics model $inv$.

\[
L_{*}^{g} = -\frac{1}{K} \sum_{i=1}^{K}g^{label}\ln{g}+(1-g^{label})\ln(1-g)
\]

The total loss is the weighted sum of all losses, and $\lambda^{pos}$, $\lambda^{g}$ are the hyper-parameters.
\[L_{total} = \lambda^{pos} (L_{act}^{pos}+L_{inv}^{pos}) + \lambda^{g} (L_{act}^{g}+L_{inv}^{g})\]

\section{Experiments}
The goal of our experiments is to verify following assumptions: (1) the novel \textit{stage conscious attention} has the ability to locate each primitive (stage) in \textbf{length-variant demonstrations} and highlight \textbf{important frames} for each playout frame. (2) SCAN can learn a \textbf{relationship mapping} between different types of experts and agent. (3) Based on above assumptions, SCAN can retrieve knowledge from few demonstrations simultaneously and get a boosted performance rather than separately handling each demonstration. 

\paragraph{Experiment Settings.} We have two main experiments. (1) we evaluate all methods on two compound tasks, \textit{pick \& place (PP)} and \textit{pick \& place \& push (PPP)}, as shown in Figure \ref{fig:env}.  In \textit{PP task (2 stages)}, the agent needs to pick the cube and place it in the target bowl. Another bowl serves as the disruptor. Next, in \textit{PPP task (3 stages)}, the agent needs to push the sky-blue cup off the table after placing the cube. Note that the target bowl might be the front one or rear one. This means the directions of the trajectory are quite different, which makes our experiment challenging. Moreover, we follow the evaluation protocol in FSL problem. There are 56 novel environments composed of unseen objects. For each environment, we let methods play 20 times with different initial scenes and calculate the success rate. The average of successful rate and standard deviation in all novel environments are provided in Table \ref{table:result}. This experiment aims to evaluate whether SCAN can locate important frames and achieve dominating performance. (2) We design a extremely length-biased case to observe the robustness of methods when encountering sub-optimal demonstrations (still complete the task but with trivial moves) that have not been processed. To be clear, all methods are trained with optimal demonstrations. Then, we give few sub-optimal demonstrations in testing to analyze the relation between attention mechanism and performance changes, as shown in Figure \ref{fig:extreme} and \ref{fig:extreme_performance}.  For all experiments, we build environments in CoppeliaSim and use the pyrep \cite{james2019pyrep} toolkit to communicate with environments. We use Panda arm as the agent, and experts may be Panda arm or UR5.

\begin{table*}[t]

\centering
\begin{tabular}{ccccccc}
\toprule
\multirow{2}{*}{Type} & \multirow{2}{*}{Models} & \multirow{2}{*}{Fine-tune} & \multicolumn{2}{c}{PP task}  & \multicolumn{2}{c}{PPP task} \\

& & &  1-shot & 5-shot &  1-shot & 5-shot \\
\midrule
\multirow{8}{*}{same} & \small BC &  & \multicolumn{2}{c}{01.70\% $\pm$ 03.31\%} &  \multicolumn{2}{c}{03.30\% $\pm$ 4.04\%} \\
& \small meta-BC & \checkmark & 12.86\% $\pm$ 12.74\% & 28.84\% $\pm$ 09.82\% & 05.54\% $\pm$ 09.76\% & 28.57\% $\pm$ 12.67\% \\
& \multirow{2}{*}{\small TaskEmb \shortcite{James2018}} &  & 35.09\% $\pm$ 34.99\% & 34.11\%  $\pm$ 34.50\% & 17.77\% $\pm$ 18.47\% & 18.39\% $\pm$ 19.66\% \\
& & \checkmark & 54.20\% $\pm$ 25.74\% & 83.39\% $\pm$ 13.47\% & 47.23\% $\pm$ 18.85\% & 58.39\% $\pm$ 18.49\% \\
& \multirow{2}{*}{\small TANet (ours)} &  & 28.75\% $\pm$ 14.24\% & 40.75\% $\pm$ 12.37\% & 46.43\% $\pm$ 21.69\% & 48.13\% $\pm$ 18.79\% \\
& & \checkmark & 53.93\% $\pm$ 20.54\% & 64.82\% $\pm$ 15.84\%  & 52.05\% $\pm$ 21.04\% & \textbf{68.57\% $\pm$ 14.16\%} \\
&\multirow{2}{*}{\small SCAN (ours)} &  & \textbf{67.05\% $\pm$ 21.31\%}  & 75.45\% $\pm$ 17.66\% & 46.34\% $\pm$ 13.18\% & 47.86\% $\pm$ 13.69\% \\
& &  \checkmark & 64.64\% $\pm$ 21.50\%  & \textbf{85.00\% $\pm$ 10.69\%} & \textbf{55.80\% $\pm$ 20.24\%} & 58.48\% $\pm$ 22.56\%\\
\hline
\rule{0pt}{2ex}
\multirow{4}{*}{differ} & \small meta-BC & \checkmark & 06.52\% $\pm$ 09.95\% & 05.80\% $\pm$ 08.17\% & 00.00\% $\pm$ 00.00\% & 00.00\% $\pm$ 00.00\% \\
& \small TaskEmb \shortcite{James2018} &  & 18.04\% $\pm$ 09.81\% & 18.66\% $\pm$ 09.61\% & 12.32\% $\pm$ 14.70\% & 12.23\% $\pm$ 15.12\%  \\
& \small TANet (ours) &  & 42.50\% $\pm$ 14.82\%  &  47.95\% $\pm$ 14.63\% & 24.02\% $\pm$ 26.30\% & 25.54\% $\pm$ 27.20\% \\
 & \small SCAN (ours) &  & \textbf{60.89\% $\pm$ 14.70\%}  &  \textbf{65.27\% $\pm$ 13.74\%} & \textbf{31.52\% $\pm$ 10.60\%} & \textbf{32.41\% $\pm$ 11.38\%} \\
\bottomrule
\end{tabular}
\caption{\textbf{Success rate on compound tasks.} The average of success rate and standard deviation in all novel environments are provided. The type column represents whether the expert and the agent are the same or not. In the same expert setting, model can fine-tune when expert actions are provided. From the table, we have three main findings. (1) the 5-shot performance usually outperforms the 1-shot performance. (2) fine-tuning is helpful in most cases, but it might let model overfit on the demonstration in one-shot setting. (3) SCAN has the best adaptation ability and performance except for the case of same expert in PPP task.}
\label{table:result}
\end{table*}

\paragraph{Compared Baselines.} Baselines are introduced below. All methods use our visual head and ActorNet for fair comparisons. Only the parts that handle few demonstrations are implemented. (1) \textbf{BC}: a conventional BC model takes states as input, no task-embedding generated. (2) \textbf{meta-BC}: a conventional BC is trained via MAML \cite{Finn17MAML} in same expert setting and trained via DAML \cite{Yu2018} in different expert setting. (3) \textbf{TaskEmb}: a DC method \cite{James2018} averages concatenated embeddings of first and last frames as task-embeddings.(4) \textbf{TANet}: our implemented DC method that averages the output of cross-demonstration attention (at each timestamp) and apply the global attention to get the task-embeddings. The key idea of TANet is similar to the method in \cite{dance21}, however, it is hard to build a transformer-based model with our visual head and ActorNet. Therefore, we design the TANet to evaluate the effectiveness of cross-demonstration attention. 

\subsection{Performance Comparison}
We analyze the performance results of experiment 1 in Table \ref{table:result}. The success rate and standard deviation is the average of 56 novel environments. We have several observations from the results. (1) Except for TaskEmb, methods achieve better performance in 5-shot setting rather than 1-shot setting. TaskEmb only uses frame features when generating task-embedding, and there is no other conversion process. Therefore, it is susceptible to frame features from novel environments. Without fine-tuning, TaskEmb performance of 1-shot and 5-shot is not much different. (2) The performance of DC methods has been dramatically improved after fine-tuning. But SCAN has slightly worse performance in PP task under the 1-shot setting. We infer that using only one demonstration to fine-tune may let models overfit. Therefore, the generalization of models is reduced. (3) SCAN has the best adaptability in most cases, regardless of whether the expert is the same as the agent. We claim that the proposed SCA learns the mapping from demonstration to playout. And, the learned mapping can provide enough information for SCAN to behave in a novel environment even without fine-tuning.

Furthermore, we also observe an interesting phenomenon. TANet has trouble in PP task, even with fine-tuning. Because the time misalignment between demonstrations in PP tasks is serious, averaging information at each timestamp like TANet causes the task to be incomprehensible. Our SCAN identifies the location of critical frames in demonstrations. Therefore, SCAN can learn and behave smoothly in PP task. However, TANet outperforms SCAN in the PPP task under the same expert setting. We infer two possible reasons regarding this phenomenon: (1) Although demonstrations of PPP task have longer lengths and more steps, their time misalignment are not severe. (2) Bowls are at the front of the agent, and many bowls have similar colors to the agent in novel environments, which interferes with SCAN that needs to pay attention to each frame. 

\begin{figure}[!t]
    \centering
    \begin{subfigure}[b]{.49\columnwidth}
    \centering
        \includegraphics[width=0.97\columnwidth]{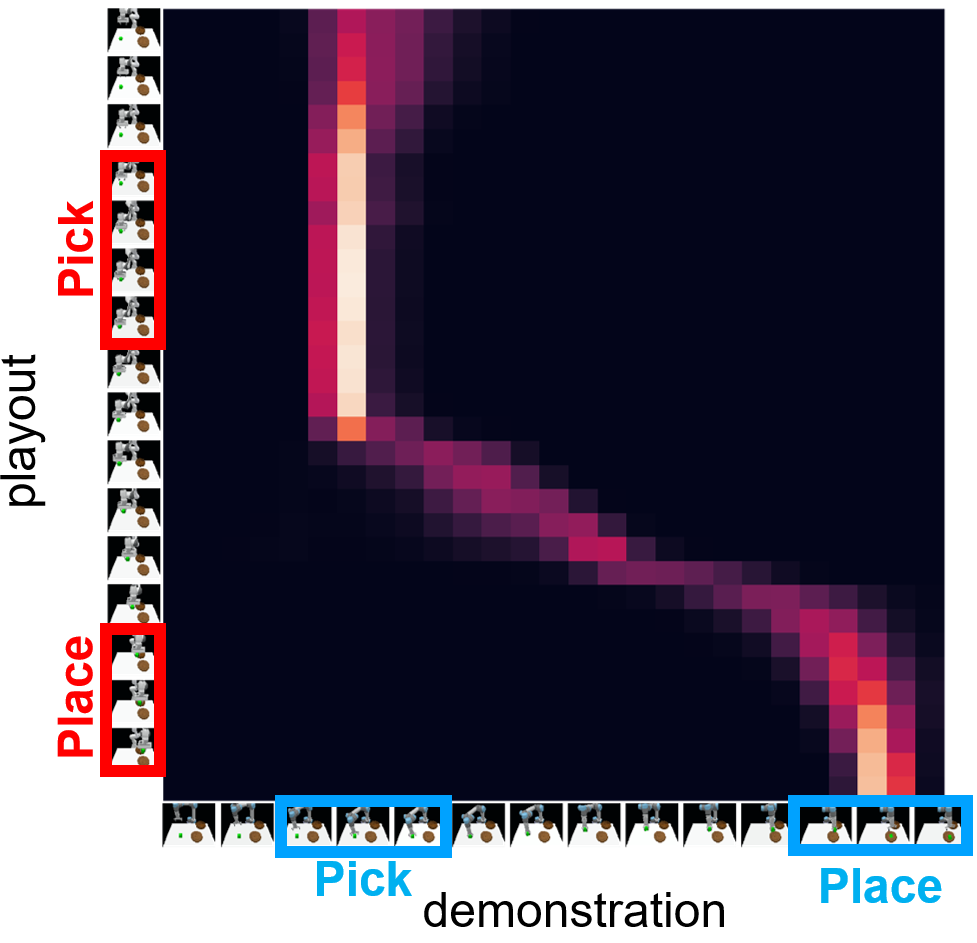}
    \caption{\textit{PP task} (2 primitives)}
    \end{subfigure}
    \begin{subfigure}[b]{.49\columnwidth}
    \centering
        \includegraphics[width=1.\textwidth]{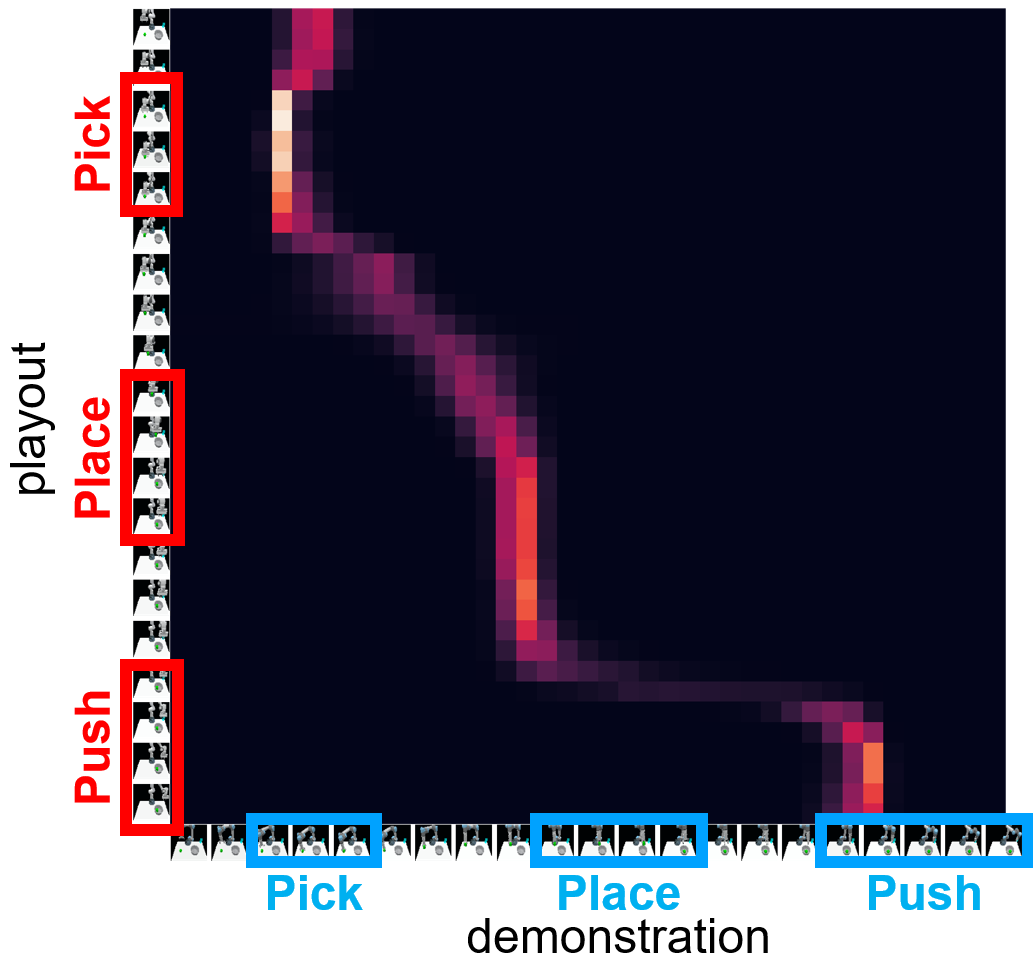}
    \caption{\textit{PPP task} (3 primitives)}
    \end{subfigure}
    \caption{\textbf{Attention maps of SCAN on compound tasks.} Each row and column represent the frame from the playout and the demonstration. Besides, a lighter cell has a higher score, and we mark each primitive with the same color. The attention results show that SCAN can focus on corresponding frames (beginning of same stage) in the demonstration when executing each stage (in both tasks).  }
    \label{fig:attention}
\end{figure}

\begin{figure}[t]
    \centering
    \begin{subfigure}[b]{.49\columnwidth}
    \centering
        \includegraphics[width=0.95\columnwidth]{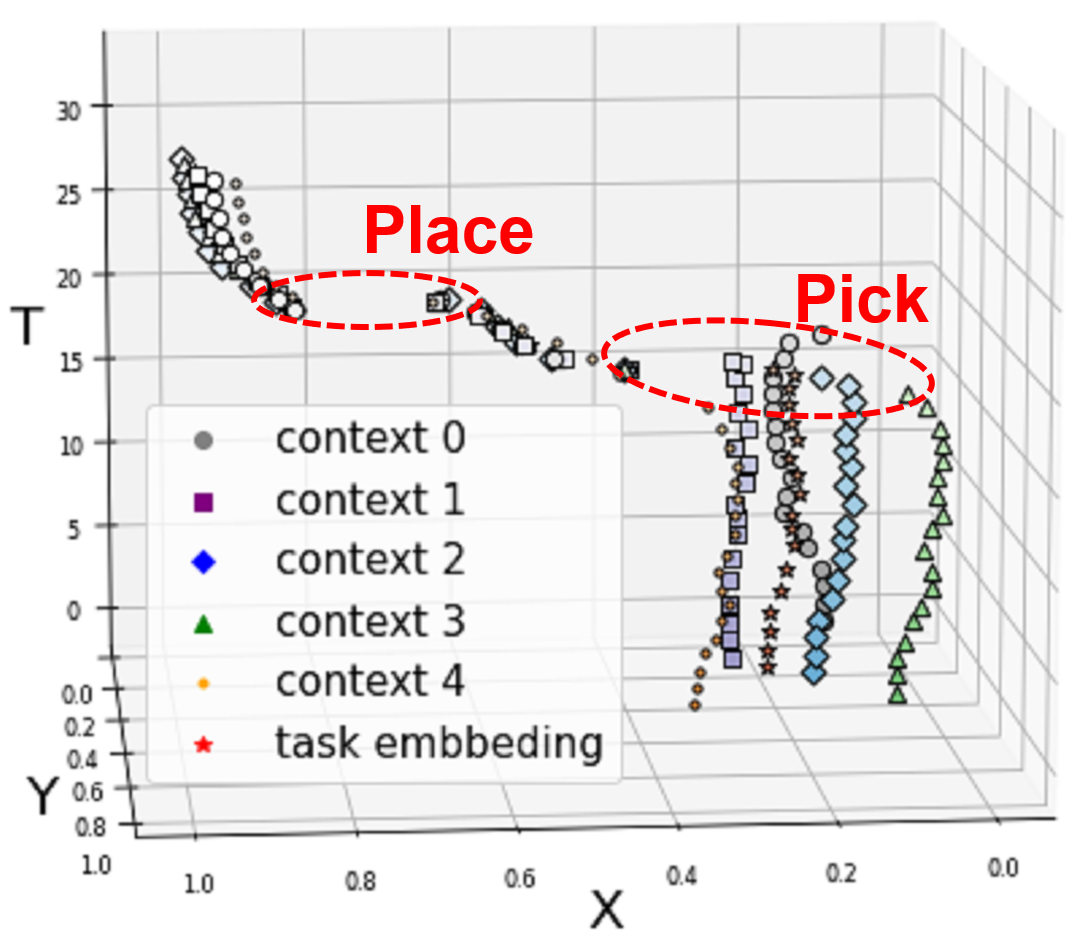}
    \caption{\textit{PP task} (2 primitives)}
    \end{subfigure}
    \begin{subfigure}[b]{.49\columnwidth}
    \centering
        \includegraphics[width=01.\columnwidth]{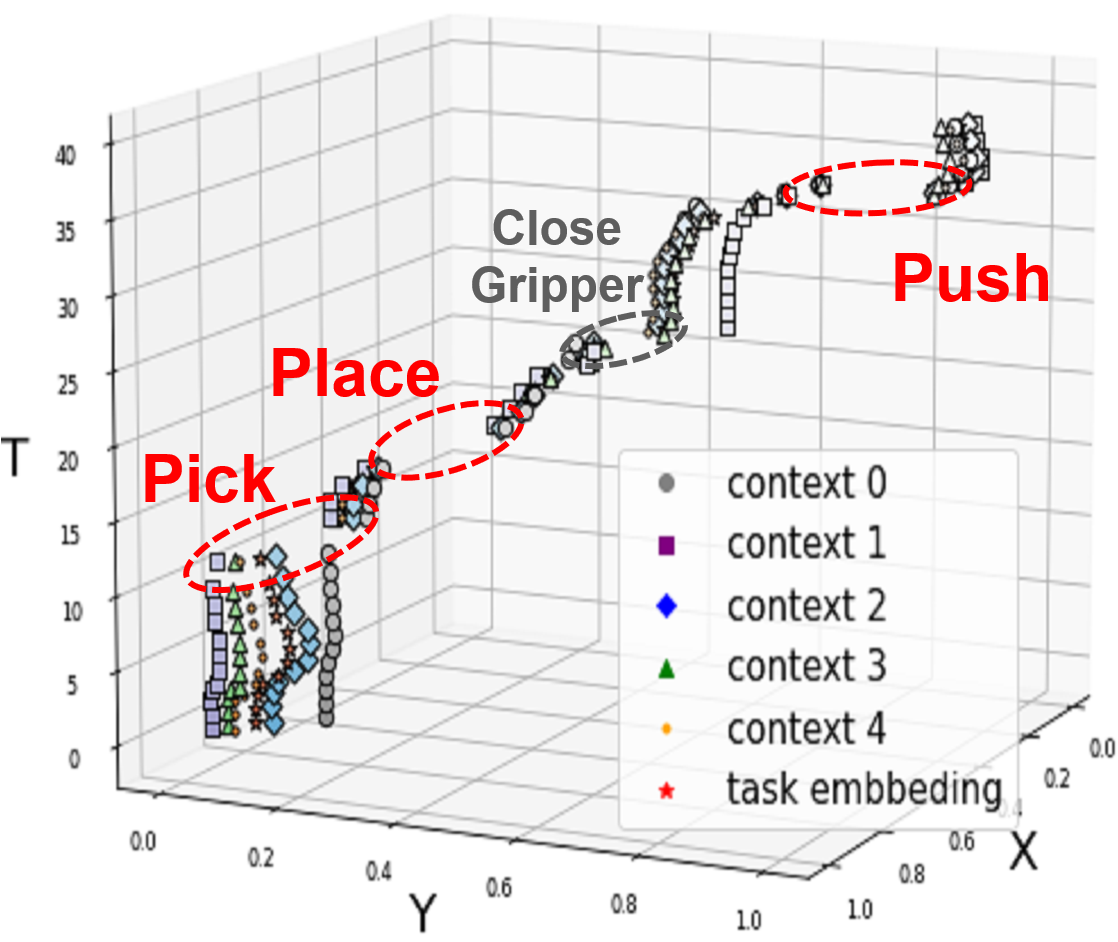}
    \caption{\textit{PPP task} (3 primitives)}
    \end{subfigure}
    \caption{\textbf{t-SNE of SCAN contexts.} The X, Y axes are the projection of t-SNE, and the T-axis represents the timestamp. A context at time $t$ (datapoint at $(x, y, t)$) is generated by time $t$ playout frame and a demonstration. Besides, contexts from different demonstrations are tagged with distinct marker. We illustrate how the relationship between contexts changes over time. Since initial states of demonstrations are various, the contexts are diverse. Surprisingly, contexts aggregate after primitives are executed, which implies that the contexts indeed contain the task-related knowledge.}
    \label{fig:tsne}
\end{figure}

\subsection{Effectiveness of Stage Conscious Attention (SCA)}
\paragraph{Attention Result in Compound Tasks.} To verify whether SCA can detect crucial frames and generate informative task-embedding, Figure \ref{fig:attention} and Figure \ref{fig:tsne} visualize attention maps and contexts generated by SCA in the compound tasks of experiment 1. To emphasize, all visualizations are under the setting of 5-shot and different experts. In Figure \ref{fig:attention}, attention scores of the stage where the agent is in progress focus on the same stage in demonstrations, whether the compound task has 2 or 3 stages. Notably, SCA locate critical frames in novel environments without fine-tuning. Furthermore, we do not use any hard restriction or loss function to guide SCA. It learns the ability proactively. On the other hand, Figure \ref{fig:tsne} illustrates t-SNE \cite{vandermaaten08a} results of contexts and task-embeddings in the playout of Figure \ref{fig:attention}. The contexts come from different demonstration are tagged with distinct marker. In addition, we highlight stage locations in t-SNE results. At the beginning, generated contexts and task-embeddings are diverse since the initial states of demonstrations are various. Specially, contexts aggregate when each stage is over. The rest of contexts are generated during moving, such as moving to target cubes or bowls. It is impressive that contexts have these cluster attributes.

\begin{figure}[t]
    \centering
    \includegraphics[width=1.\columnwidth]{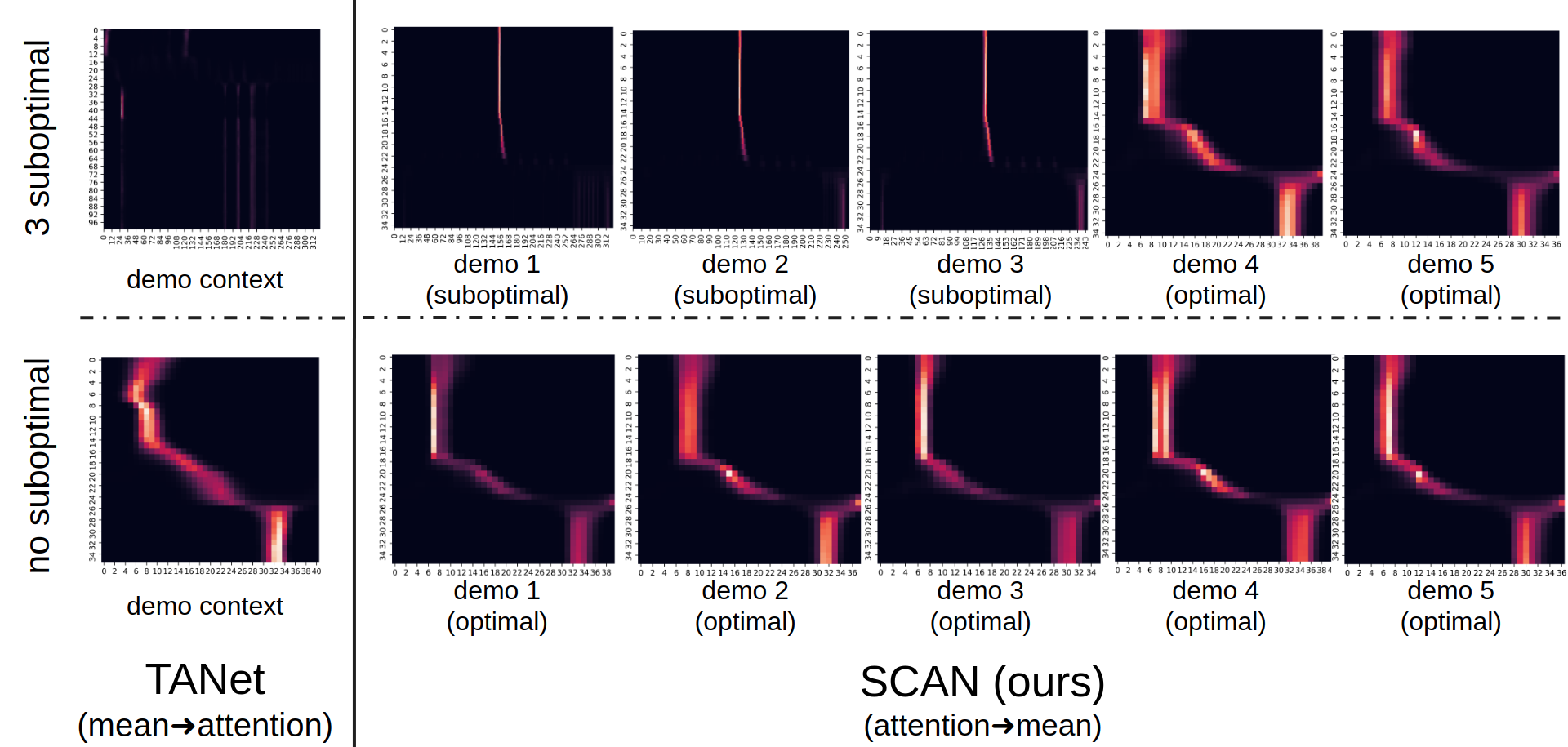}
    \caption{\textbf{Attention in extremely length-biased case.} Methods are trained with optimal demonstrations, but the demonstration set contains sub-optimal demonstrations (with trivial moves) in testing. When there are all optimal demonstrations, both TANet and SCAN can find important frames. However, the attention result of TANet becomes a mess when there are 3 sub-optimal demonstrations. Because TANet averages attention by each timestamp, the information are mixed and hard to retrieve. By contrast, SCAN generates attention to each demonstration individually and then averages the results. Since unimportant frames has been filtered out, SCAN can still extract important information.}
    \label{fig:extreme}
\end{figure}

\begin{figure}[!t]
    \centering
    \includegraphics[width=0.9\columnwidth]{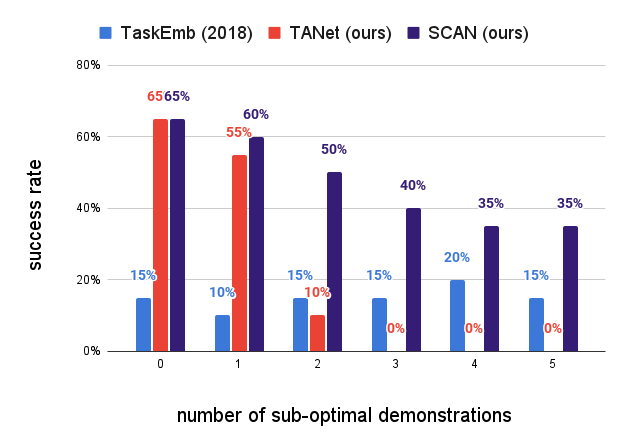}
    \caption{\textbf{Model robustness in extremely length-biased case.} We analyze the relation between success rate and the number of sub-optimal demonstrations. TANet cannot handle sub-optimal demonstrations since they average attention by timestamp. Its performance dramatically decreases while the number of sub-optimal demonstrations increases. In the contrary, our SCAN is more robust to the demonstrations with extremely different length. TaskEmb only considers the first and last frame, therefore, it does not matter in this case.    }
    \label{fig:extreme_performance}
\end{figure}

\paragraph{Robustness to Sub-optimal Demonstrations.} Figure \ref{fig:extreme} and Figure \ref{fig:extreme_performance} demonstrate the result of experiment 2. We choose one from 56 novel environments as the target environment and generate the sub-optimal demonstrations. Experts still completed the task in sub-optimal demonstrations but with trivial movements. In other words, the demonstration set is extremely length-biased, which all methods have never encountered during training. Then, we run all methods 20 times for the performance evaluation. Attention maps of SCAN and TANet baseline in this experiment are shown in Figure \ref{fig:extreme}. It is the comparison with zero or three sub-optimal demonstrations in the demonstration set. Both methods work well when there are no sub-optimal demonstrations (bottom). However, TANet cannot handle the case that there are three sub-optimal demonstrations (top). Because the time misalignment is severe in the case, TANet is hard to retrieve knowledge by averaging at each timestamp of demonstrations, which can be reflected in success rates of Figure \ref{fig:extreme_performance}. 

In Figure \ref{fig:extreme_performance}, TANet performs poorly when the number of sub-optimal demonstrations increases. Our SCAN is also affected by sub-optimal demonstrations, but it did not cause such a big reduction in performance. Moreover, TaskEmb only focuses on first and last frames of demonstrations, and thus, sub-optimal demonstrations would not affect its performance. However, first and last frames cannot provide efficient information when solving compound tasks, TaskEmb has a lower performance compared to other methods.


\section{Conclusion}
In this work, we conduct the FSIL problem under three challenging settings, including compound tasks, few length-variant demonstrations and learning from a different expert. Meanwhile, we found that most of works can only handle one demonstration at once or need external loss to learn from different experts. Hence, we propose a novel SCAN method that can retrieve knowledge from few demonstrations simultaneously and behave in novel environments without fine-tuning. Our stage conscious attention locates critical frames for each playout frame to alleviate the demonstration misalignment problem. Explainable visualization and outstanding performance illustrates the effectiveness of SCAN.

\section{Acknowledgments}
This work was supported in part by the Ministry of Science and Technology, Taiwan, under Grant MOST 110-2634-F-002-051 and Qualcomm Technologies, Inc. We are grateful to the National Center for High-performance Computing.

\bibliography{ref.bib}


\section*{Supplementary material}

# Stage Conscious Attention Network (SCAN) :
# A Demonstration-Conditioned Policy for Few-Shot Imitation
# Supplementary Material

The supplementary material contains following contents:

- Summary of previous few-shot imitation methods
- Comparison between the conventional policy and the demonstration-conditioned policy.
- Details of proposed SCAN, including the architecture of visual head and implementation details.
- The progress how we build environments and collect demonstrations.
- Experiment on the importance of designed components.
- Verification and visualization of stage conscious attention (SCA) in different cases.

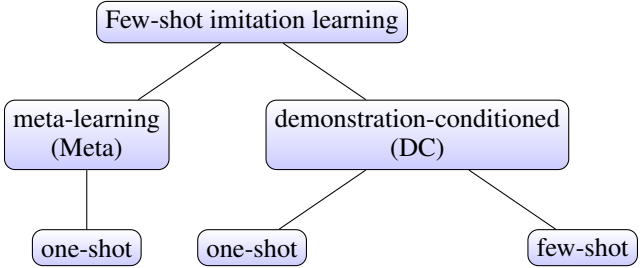

Figure 1: Previous FSIL works are categorized into meta-learning and demonstration-conditioned based. Meta-learning methods seek a meta-parameter set that can easily adapt to novel tasks. Besides, demonstration-conditioned methods generate actions by using both current states and demonstrations as input. Then, we classify methods into one-shot or few-shot based on how they use demonstrations to update their parameters. Only works that process few demonstrations simultaneously belong to few-shot class. Thus, meta-learning methods can only be in one-shot class.

## Few-Shot Imitation Learning Methods

To provide a whole picture of FSIL methods, we list all classes of FSIL methods in Figure 1. Moreover, we summarize them according to their architecture and problem setting in Table 1. As illustrated in Table 1, we are the first work that

---

**Algorithm 1:** Meta-train DC policy with $K$-shot demo

**Require**: *base* environments $E^b$

1: $i = 1, \theta = RandomInit()$
2: **while** $i < MaxEpoch$ **do**
3:   **for** each $e^b$ in $E^b$ **do**
4:     $j = 1, \mathcal{D}^b = \{\}$
5:     **while** $j \leq K$ **do**
6:       Sample a video-only demo $d^b := \{\mathbb{F}\}^b$ from $e^b$
7:       $\mathcal{D}^b = \mathcal{D}^b \cup \{d^b\}$
8:       $j = j + 1$
9:     **end while**
10:     Sample a playout $p^b := (\mathbb{S}, \mathbb{A})^b$ from $e^b$
11:     $\hat{\mathbb{A}} = \pi_\theta(\mathbb{S}, \mathcal{D}^b)$
12:     $\theta = \theta - \nabla_\theta \mathcal{L}_{BC}(\theta, \mathbb{A}, \hat{\mathbb{A}})$
13:   **end for**
14:   $i = i + 1$
15: **end while**
16: **return** $\theta$

---

conduct FSIL problem on all challenge settings. In following paragraphs, we introduce the demonstration-conditioned policy and compare it with conventional methods.

## Demonstration-Conditioned Policy

The demonstration-conditioned (DC) policy $\pi_\theta(a|s, \mathcal{D}^*)$ needs both current observation $s$ and demonstrations $D^*$ as inputs to generate actions. We compare the conventional policy and demonstration-conditioned policy in Figure 2. When using conventional policy, demonstrations are served as batch data. The conventional policy fine-tunes its parameters with demonstrations and then discards them. They would not access the demonstrations when generating actions for current playout $p^n$. This design has several limitations. (1) The demonstrations must contain expert actions for the supervised fine-tuning. (2) the policy might be misled if the expert in demonstrations is different from the agent.

By contrast, demonstration-conditioned policy treats demonstrations as the support set and learns the relationship between demonstrations and playouts during training. In addition, due to the characteristic of FSIL problem, the DC policy is equivalent to the metric-learning-based method in FSL. They both learn the relationship rather than infer-

| Works | Type | C | F | D |
|---|---|---|---|---|
| Duan et al. (2017) | DC | ✓ | | |
| Finn et al. (2017) | Meta | | | |
| Yu et al. (2018) | Meta | | | ✓ |
| James, Bloesch, and Davison (2018) | DC | | ✓ | |
| Yu et al. (2019) | Meta | ✓ | | ✓ |
| Bonardi, James, and Davison (2020) | DC | | ✓ | ✓ |
| Dasari and Gupta (2020) | DC | | | ✓ |
| Cachet, Perez, and Kim (2020) | Meta | | | |
| Dance, Perez, and Cachet (2021) | DC | | ✓ | |
| SCAN (Ours) | DC | ✓ | ✓ | ✓ |

Table 1: **Problem setting of existing FSIL works.** Our FSIL problem is under three challenging settings, including compound tasks ($\geq 3$ stages) **(C)**, learn from few length-variant and misalignment demonstrations concurrently **(F)**, and learn the behavior from a different type of demonstrator **(D)**. We summarize existing FSIL methods according to algorithm architectures and problem settings. Duan et al. (2017) is the first FSIL work. They focus on the stacking task which contains multiple times of pick and place stages. Then, Finn et al. (2017) introduces the meta imitation learning (MIL). Afterwards, Yu et al. (2018) proposes the domain adaption meta-learning (DAML) which allows MIL algorithm imitates from different types of experts. Besides, James, Bloesch, and Davison (2018) is the first work based on the concept of task-embedding in FSIL problem. And them (2020) further extends it to imitate from a human demonstrator. Yu et al. (2019) applies DAML on compound tasks. Both Dasari and Gupta (2020) and Cachet, Perez, and Kim (2020) propose transformer-based methods to solve FSIL problem. At last, Dance, Perez, and Cachet (2021) extends their method to support few demonstrations.

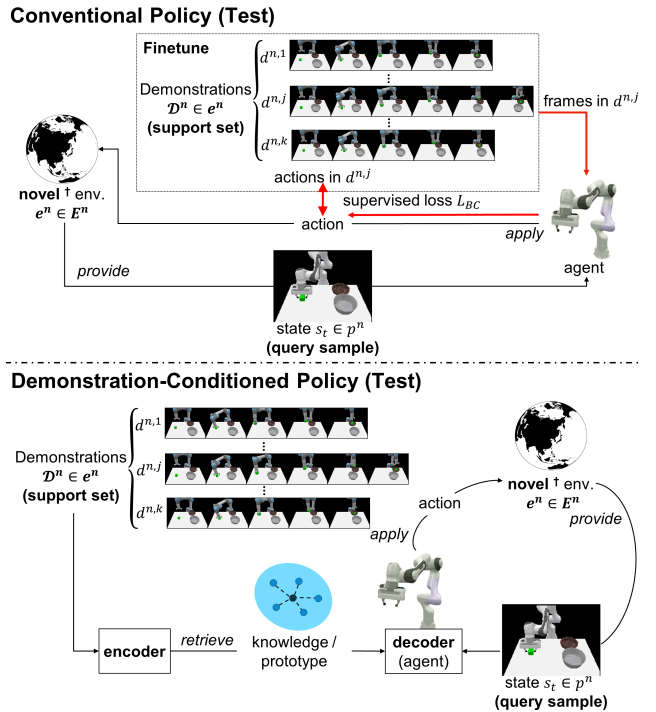

Figure 2: **Comparison of conventional policy and demonstration-conditioned policy.** Conventional policy is usually be trained and tested in the same environments. When solving FSIL problem, it needs further fine-tuning in novel environments $e^n$. It means that demonstrations $\mathcal{D}^n$ must contains actions for the supervised learning. After fine-tuning, conventional policy does not access demonstrations when generating actions for current playout $p^n$. By contrast, the fine-tuning is optional for DC policy. Since they are trained to learn the relationship between demonstrations and current playout, they retrieve knowledge from demonstrations and directly compute actions for the current state $s_t$ in the novel environment $e^n$. The benefit of DC policy is that its performance is more robust when the expert and agent are different. Using demonstrations to fine-tune might mislead the conventional policy model, since the agent is not in the demonstrations.

ring the data directly. A similar finding is also discussed in (James, Bloesch, and Davison 2018). Fine-tuning is optional for DC policies since they are not learning how to behave in a specific environment. Furthermore, most of them have an encoder-decoder architecture. Demonstrations are fed into the encoder for knowledge retrieval, and the decoder generates actions according to the extracted knowledge and current state. The pseudocode of training and testing a DC policy is in Algorithm 1 and Algorithm 2, respectively.

DC policy is widely used in FSIL works, and they use various ways to process the demonstrations. However, these methods, such as feeding a whole demonstration (Dasari and Gupta 2020), concatenating first and last frames (James, Bloesch, and Davison 2018), or leveraging a cross-demonstration attention (temporal-awareness attention) and assuming each timestamp of different demonstrations contains similar information, cannot effectively work on FSIL problem with our settings. Thus, we aim to develop a method that highlights critical frames for each playout states.

## Stage Conscious Attention Network (SCAN)

In this section, we describe the details of SCAN not mentioned in the main paper. First, we introduce the modified self-attention module and architecture of visual head. Then, the implementation details such as used optimizer, learning rate adjustment, and fine-tuning progress are provided.

### Details of Visual Head

**Self-Attention Module.** Some works (Shao et al. 2020) have to use the object detection model to segment objects in visual inputs before generating actions. However, such a model is hard to train all its components from scratch. Inspired by (Wang et al. 2017), we assume that adding attention modules in feature extractor to extract more informative features may be an alternative method. Therefore, as shown in the top and mid of Figure 3, we introduce an attention version of ResNet18 (He et al. 2016), denoted as SA-ResNet18. To be specific, we insert a modified self-attention

Algorithm 2: Meta-test DC policy with $K$-shot demo
**Require**: *novel* environments $E^n$
**Require**: trained parameters $\theta$
**Require**: number of playout $N^{ply}$
1: **for** each $e^n$ in $E^n$ **do**
2:   $j = 1, \mathcal{D}^n = \{\}$
3:   **while** $j \leq K$ **do**
4:     Sample a video-only demo $d^n := \{\mathbb{F}\}^n$ from $e^n$
5:     $\mathcal{D}^n = \mathcal{D}^n \cup \{d^n\}$
6:     $j = j + 1$
7:   **end while**
8:   $k = 0, cnt = 0$
9:   **for** $k \leq N^{ply}$ **do**
10:    $e^n = RandomReset()$
11:    $s_0 = initial\ state$ from $e^n$
12:    $t = 0, \mathbb{S} = [s_0]$
13:    $done = False$
14:    **while** not $done$ **do**
15:      $\hat{\mathbb{A}} = \pi_\theta(\mathbb{S}, \mathcal{D}^n)$
16:      $s_{t+1}, done = e^n(s_t, a_t)$
17:      $\mathbb{S} = \mathbb{S}.append(s_{t+1})$
18:      $t = t + 1$
19:    **end while**
20:    **if** success **then**
21:      $cnt = cnt + 1$
22:    **end if**
23:    $k = k + 1$
24:   **end for**
25:   success rate $sr = cnt/N^{ply}$
26: **end for**
27: calculate average of $sr$ and $\sigma$ over all $e^n$ in $E^n$

module at the head and tail of the ResNet18. Besides, the self-attention module is originally proposed in (Zhang et al. 2019). We change the output dimension of $1 \times 1$ convolution layer (query-layer and key-layer) from 1 to the input dimension, and then a channel-wise softmax layer is followed.

The motivation for this modification is that the input dimension of the self-attention module might be huge, such as 256 or 512. Using a 1x1 convolution layer to reduce such high-dimensional features to one dimension may cause loss of information. Hence, we let the output of these layer has the same dimension as the input and use the channel-wise softmax to decide the importance of each dimension in the output. The modified self-attention module can provide more meaningful knowledge since it retrieves and reserve the knowledge from input features effectively.

**Architecture of Visual Head.** As mentioned in Figure 3, visual head uses two SA-RestNet18 to extract RGB and depth images separately. This design is motivated by (Shao et al. 2020), and the intuition is that self-attention modules in SA-ResNet18 may focus on distinct areas of RGB and depth images. Then, we concatenate the output features and feed them into a fully connected layer followed by a dropout function with a dropout rate of 0.5. Since we deal with few-shot demonstrations, the dropout layer can avoid the following network components from overfitting on these data,

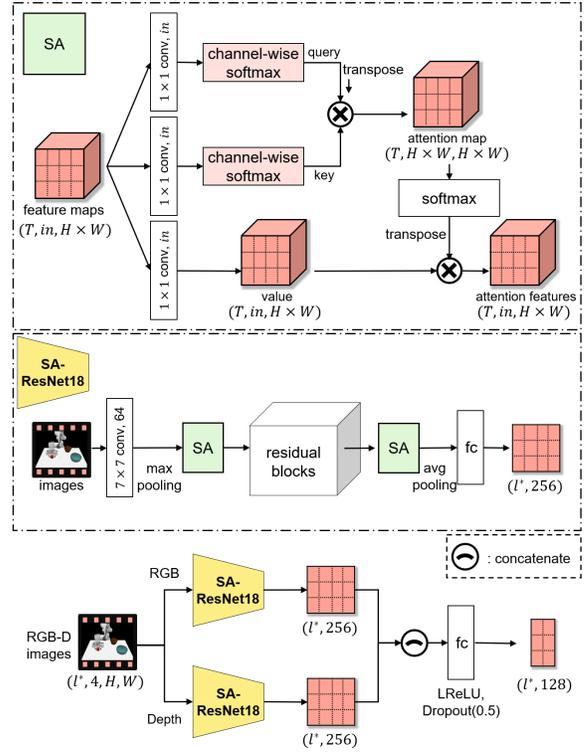

Figure 3: **Architecture of visual head.** At the top, we modify the self-attention module by changing the query-layer and key-layer output dimension and adding a channel-wise softmax behind the layers. Using a channel-wise attention can avoid the loss of information. In addition, there are two self-attention modules inserted at the head and tail of SA-ResNet18 (mid). SA-ResNet18 recognizes important information without the assistance of object detection models. At last, the visual head (bottom) uses two separate SA-ResNet18 to extract feature embeddings of RGB images and depth images. Since these two SA-ResNet18 may focus on different areas of RGB and depth images, this design can produce more meaningful feature embeddings.

which is beneficial to learn a more general model.

### Implementation Details

In this section, we provide implementation details not stated in the main paper. Hyper-parameters for Visual head, Actor-Net and training/testing are listed in Table 2 - 4. The dimension of output embeddings from visual head is 128. All activation functions are LeakyReLU except for the last hidden layer of ActorNet. We use Tanh function to limit the position outputs to a specified range and apply Softmax function to generate the probabilities of open/close control.

All methods are trained with ten epochs. There are 110 base environments in an epoch. For each base environment, five demonstrations and five playouts are sampled for training the models. No interactions with environments are allowed during training. Besides, model parameters are optimized by an Adam optimizer with the learning rate $2 \times 10^{-4}$.

| Hyper-parameter | Value |
|---|---|
| Dimension of output embedding | 128 |
| Activation function | LeakyReLU |
| Dropout rate | 0.5 |

Table 2: Hyper-parameters for Visual head

| Hyper-parameter | Value |
|---|---|
| Dimension of output embedding | 128 |
| Activation for hidden layer | LeakyReLU |
| Activation for position layer | Tanh |
| Activation for open/close layer | Softmax |
| Max length of PositionalEncoding | 100 |

Table 3: Hyper-parameters for ActorNet

| Hyper-parameter | Value |
|---|---|
| Number of epochs for training | 10 |
| Number of demonstrations for training | 5 |
| Number of playouts for training | 5 |
| Number of trails for testing | 20 |
| Max time step for each trail in testing | 100 |

Table 4: Hyper-parameters used for training/testing

If fine-tuning is applied, the learning rate for fine-tuning is $2 \times 10^{-5}$. The final value of hyper-parameters is determined after several manual adjustments. We implement methods in PyTorch and conduct all experiments on a Ubuntu system that contains an Intel i9-9900KF CPU, 64GB RAM, and an NVIDIA RTX3090 24GB GPU. The approximate training time is two to three hours, and the testing time is three to five hours, depending on the performance of methods.

# Environment Details

## Composition of Base and Novel Environments

As stated in main paper, we build environments in CoppeliaSim, a robotic simulated software. About target objects that the agent need to pick, we use the cube object provided in CoppeliaSim and set its color into different colors during training and testing. Next, we download free 3D models of bowls from the internet. As shown in Figure 4, we collect eleven bowls for training (base environments) and eight bowls for testing (novel environments). Each bowl is paired with other bowls to form a environments. Thus, we have 110 base environments and 56 novel environments for each task.

## Settings of Compound Tasks

We have two compound tasks in experiment 1. The operation space for both tasks is a table of 60 cm by 75 cm. Figure 5 illustrates the placement area of objects and the movement direction of the agent. All objects are set to dynamic mode, which means the objects may change positions or even leave the operating space due to collisions. Moreover, our compound tasks are challenging since the movement direction varies in different novel environments, and we do not apply fine-tuning to let the model adapt to the trajectory. Besides,

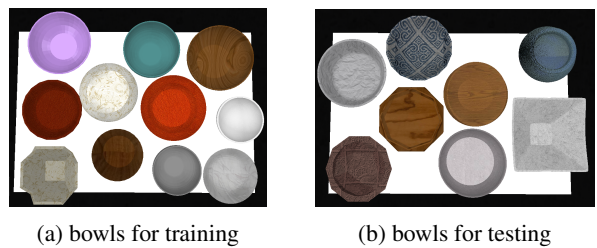

(a) bowls for training  (b) bowls for testing

Figure 4: **Bowls used in environments.** Each bowl is paired with other bowls to form an environment. (a) We have $11 \times 10 = 110$ base environments. And (b) we build $8 \times 7 = 56$ novel environments. Note that bowls in base and novel environments are disjoint. When solving a task (e.g. PP task), we train methods in base environments and evaluate their performance in novel environments.

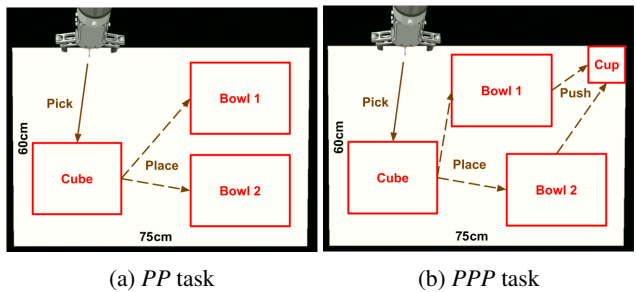

(a) *PP* task  (b) *PPP* task

Figure 5: **Configuration of compound tasks.** The red area indicates the range in which the corresponding objects may be initially placed. The operating space of the agent is a table of 60 cm by 75 cm, which means the agent needs to move a long distance to complete the task. Moreover, the target bowl (in different environments) may be Bowl 1 or Bowl 2 when placing. The directions to areas of these two bowls are quite distinct, which makes our tasks challenging.

the agent has 20 trails in each novel environment. Both positions of the cube and bowls are not fixed in each trial. In some previous works (e.g., Dasari and Gupta), only one of the positions may changes. Since our task has many challenging settings, we only have one cube on the table for the agent to pick. We would let the agent detect which object to be grasped from multiple objects in future research.

## Details of Demonstration Collection

We assume that positions of objects and the status of robotic arms are known during demonstration collection. The rule-based methods are applied to demonstrate how to solve the task. There are two robotic arms in our dataset, UR5 and Panda. For each environment, we collect 40 demonstrations of UR5 arm and 40 demonstrations of Panda arm. Each demonstration contains frames, cropped frames, end-effector vectors, and actions. With these demonstrations, user can easily switch the roles of expert and agent. The size of demonstration files in PP tasks is 500 GB and the size of demonstrations in PPP tasks is 700 GB. The duration for collecting all demonstrations is about one day.

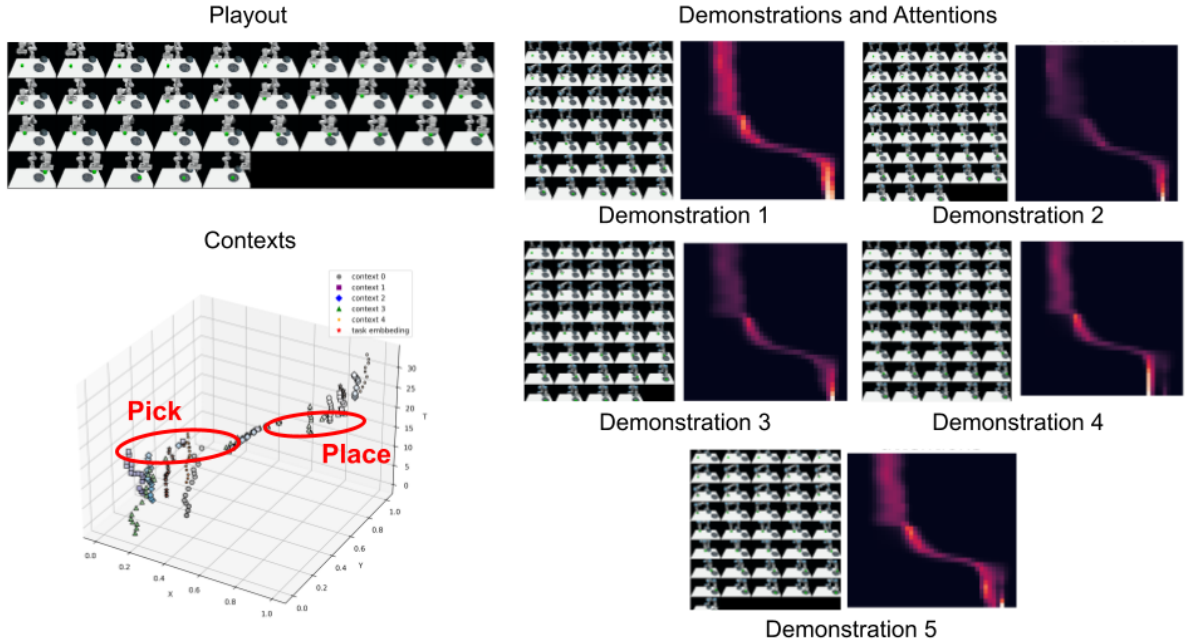

Figure 6: **Visualization of contexts and attention in a successful case.** The agent takes 35 steps to complete the task. We demonstrate playout frames of each time step in Playout (left up). In the Contexts (left bottom), the t-SNE result of contexts is generated at each time step. As previously stated, they are diverse at the beginning and converge at the end of each stage. This indicates that our stage conscious attention locate critical frames for playout frames at each time step, which can also be observed in attention results (right). In the attention results, the highest attention score of each playout frame is the frame at the same stage in the demonstrations. With the help of stage conscious attention, our method can retrieve knowledge from few misalignment demonstrations simultaneously and generate informative task-embeddings.

| Method | inv | self-attn | success rate |
|---|---|---|---|
| SCAN | ✓ | | 12.41% ± 13.98% |
| | | ✓ | 31.96% ± 17.87% |
| | ✓ | ✓ | **65.27% ± 13.74%** |

Table 5: **Importance of designed components.** The **inv** represents the inverse dynamics model, and the **self-attn** indicates the self-attention module in visual head. The goal of the inverse model is to let the model know how the action affects the changes of frames. Moreover, the self-attention module aims to assist visual head to extract informative features. When any component is removed, the performance of SCAN is significantly reduced.

## Ablation Study

To investigate the impact of components on the model, we conduct the ablation study on PP task under the setting of different expert and 5-shot and provide results in Table 5. When removing these components (inverse dynamics model and self-attention model), SCAN cannot locate the specific position of objects in novel environments, which causes a dramatically performance drop. Without the self-attention module, extracted features are not representative enough. Thus, the inverse dynamics module cannot learns the relationship between actions and the change of frames. There is a similar situation when the inverse dynamics module is removed, which proves that these components are beneficial.

## Cases Studies of Context and Attention

We are curious about how stage conscious attention (SCA) works in different situations. Therefore, Figure 6 and Figure 7 visualize three different cases when SCAN solves the PP task under the setting of different expert and 5-shot.

From Figure 6, SCAN successfully solve the PP task. Attention maps indicate the SCA can locate corresponding frames in each stage. We also observe that generated contexts are aggregated at the end of each stage. Furthermore, there are two failure cases in Figure 7, pick failure and place failure. When SCAN fails to pick the cube, the SCA focuses on the pick stage in demonstrations until the end, representing it knows the stage has not ended. Because SCAN did not complete any stages in this trial, there is no aggregation in Contexts. Then, what if SCAN succeeded in picking the cube but failed to place it? Attention results demonstrate that it focuses on frames of place stages until the end, which means SCA knows it has completed the pick stage. The Contexts in this case only contain one aggregation. The above visualization results prove that our novel SCA can extract informative knowledge and assist in solving the task.

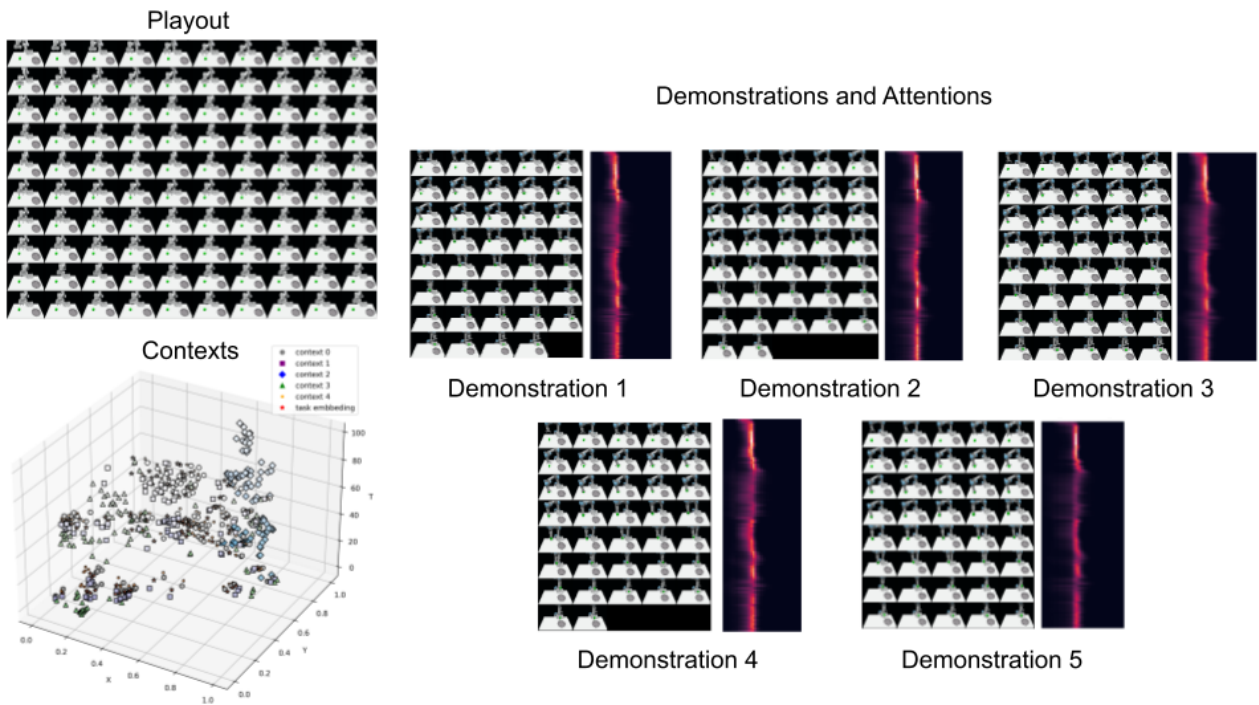

(a) pick failure

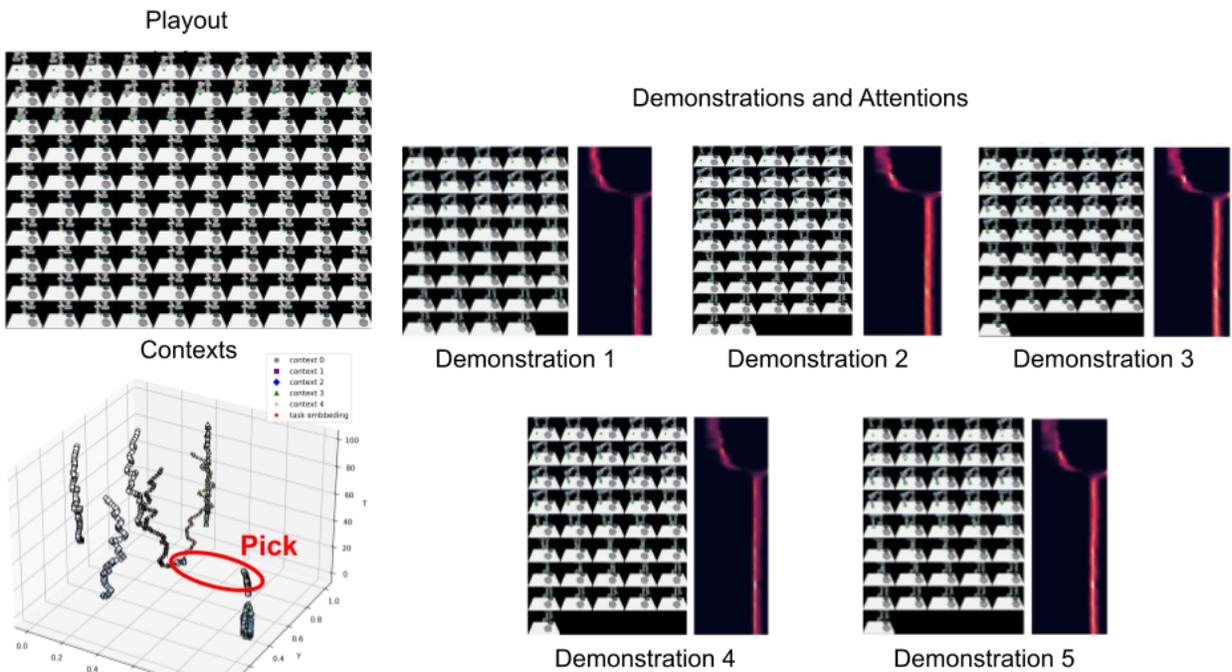

(b) place failure

Figure 7: **Visualization of contexts and attention in two different failure cases.** The max steps in a trial are 100. In (a), SCAN cannot pick the cube. We can find that SCAN has been focusing on the pick stage in demonstrations until the end, which means it knows that it has not completed the goal of this stage. Contexts are a mess in the case, and no aggregation occurs. Moreover, SCAN successfully picked the cube but did not put it in the target bowl in (b). And thus, attention scores focus on the place stage in demonstrations until the end. From the contexts, we can find an aggregation when the pick stage is completed, but there is no aggregation in the contexts generated afterward.



\end{document}